\documentclass[10pt,twocolumn,letterpaper]{article}
\usepackage[accsupp]{axessibility} 
\usepackage{wacv}
\usepackage{times}
\usepackage{epsfig}
\usepackage{graphicx}
\usepackage{amsmath}
\usepackage{amssymb}
\usepackage[accsupp]{axessibility}  
\usepackage{multirow, booktabs}
\usepackage{xcolor}         
\usepackage{colortbl}
\usepackage{amsfonts}       
\usepackage{nicefrac}       
\usepackage{microtype}      
\usepackage{xcolor}         

\usepackage{dirtytalk}
\usepackage{mathtools}
\usepackage{placeins}
\usepackage{enumerate}
\usepackage{pifont}
\usepackage{amsmath,amsfonts,bm}

\def\eqref#1{equation~\ref{#1}}

\def\ceil#1{\lceil #1 \rceil}

\def\1{\bm{1}}

\DeclareMathAlphabet{\mathsfit}{\encodingdefault}{\sfdefault}{m}{sl}
\SetMathAlphabet{\mathsfit}{bold}{\encodingdefault}{\sfdefault}{bx}{n}

\def\sN{{\mathbb{N}}}

\usepackage{enumitem}

\def\wacvPaperID{1232} 

\wacvalgorithmstrack   

\wacvfinalcopy 

\ifwacvfinal
\usepackage[breaklinks=true,bookmarks=false]{hyperref}
\else
\usepackage[pagebackref=true,breaklinks=true,colorlinks,bookmarks=false]{hyperref}
\fi

\newcommand*\samethanks[1][\value{footnote}]{\footnotemark[#1]}

\begin{document}

\title{Pushing the Efficiency Limit Using Structured Sparse Convolutions}

\author{Vinay Kumar Verma$^{1}$\thanks{The authors contributed equally to this work. Correspondence to {\tt \{vinayugc, nikhilmehta.dce\}@gmail.com}.}\enspace, Nikhil Mehta$^{1}$\samethanks \enspace, Shijing Si$^{3}$, Ricardo Henao$^{1}$, Lawrence Carin$^{1,2}$\\
$^{1}$Duke University \quad $^{2}$KAUST Saudi Arabia \quad $^{3}$SEF, Shanghai International Studies University\\
}

\maketitle

\thispagestyle{empty}

\begin{abstract}
	Weight pruning is among the most popular approaches for compressing deep convolutional neural networks. Recent work suggests that in a randomly initialized deep neural network, there exist sparse subnetworks that achieve performance comparable to the original network. Unfortunately, finding these subnetworks involves iterative stages of training and pruning, which can be computationally expensive. We propose Structured Sparse Convolution (SSC), that leverages the inherent structure in images to reduce the parameters in the convolutional filter. This leads to improved efficiency of convolutional architectures compared to existing methods that perform pruning at initialization. We show that SSC is a generalization of commonly used layers (depthwise, groupwise and pointwise convolution) in ``efficient architectures.'' Extensive experiments on well-known CNN models and datasets show the effectiveness of the proposed method. Architectures based on SSC achieve state-of-the-art performance compared to baselines on CIFAR-10, CIFAR-100, Tiny-ImageNet, and ImageNet classification benchmarks. Our source code is publicly available at \href{https://github.com/vkvermaa/SSC}{https://github.com/vkvermaa/SSC}.
\end{abstract}

\vspace{-1em}
\section{Introduction}
\label{sec:intro}
\noindent Overparameterized deep neural networks (DNNs) are known to generalize well on the test data~\cite{arora2019fine,allen2019learning}. However, overparameterization increases the network size, making DNNs resource-hungry and leading to extended training and inference time. This hinders the training and deployment of DNNs on low-power devices and limits the application of DNNs in systems with strict latency requirements. Several efforts have been made to reduce the storage and computational complexity of DNNs using model compression \cite{han2015learning,yu2017compressing,verma2020network,singh2019hetconv,louizos2017bayesian,alvarez2017compression,luo2017thinet,chen2018constraint,singh2020leveraging}. Network pruning is the most popular approach for model compression. In network pruning, we compress a large neural network by pruning redundant parameters while maintaining the model performance. The pruning approaches can be divided into two categories: unstructured and structured. Unstructured pruning removes redundant connections in the kernel, leading to sparse tensors~\cite{lee2018snip,han2015deep, zhang2018systematic}. Unstructured sparsity produces sporadic connectivity in the neural architecture, causing irregular memory access~\cite{wen2016learning} that adversely impacts the acceleration in hardware platforms. On the other hand, structured pruning involves pruning parameters that follow a high-level structure ($e.g.$, pruning parameters at the filter-level~\cite{li2016pruning,luo2017thinet,ding2018auto}). Typically, structure pruning leads to practical acceleration, as the parameters are reduced while the memory access remains contiguous. Existing pruning methods typically involve a three-stage pipeline: pretraining, pruning and finetuning, where the latter two stages are carried out in multiple stages until a desired pruning ratio is achieved. While the final pruned model leads to a low inference cost, the cost to achieve the pruned architecture remains high. 

The lottery ticket hypothesis (LTH) \cite{frankle2018lottery,frankle2019stabilizing} showed that a randomly initialized overparametrized neural network contains a sub-network, referred to as the ``winning ticket,'' that when trained in isolation achieves the same test accuracy as the original network. Similar to LTH, there is compelling evidence~\cite{neyshabur2018role,neyshabur2014search,du2018gradient,du2018power,allen2019convergence,allen2019learning,singh2019hetconv} suggesting that overparameterization is not essential for high test accuracy, but is helpful to find a good initialization for the network \cite{li2018learning,zou2020gradient}. However, the procedure to find such sub-networks involves iterative pruning~\cite{frankle2018lottery} making it computationally intensive. If we know the sub-network beforehand, we can train a much smaller and efficient model with only 1-10\% of the parameters of the original network, reducing the computational cost involved during training. 

An open research question concerns how to design a sub-network without undergoing the expensive multi-stage process of training, pruning and finetuning. There have been recent attempts~\cite{lee2018snip,wang2020picking} to alleviate this issue, involving a one-time neural network pruning at initialization by solving an optimization problem for detecting and removing unimportant connections. Once the sub-network is identified, the model is trained without carrying out further pruning. This procedure of pruning only once is referred to as pruning at initialization or foresight pruning~\cite{wang2020picking}. While these methods can find an approximation to the winning ticket, they have the following limitations hindering their practical applicability: $(1)$ The initial optimization procedure still requires large memory, since the optimization process is carried out over the original overparameterized model. $(2)$ The obtained winning ticket is specific to a particular dataset on which it is approximated, $i.e.$, a network pruned using a particular dataset may not perform optimally on a different dataset. $(3)$ These pruning based methods lead to unstructured sparsity in the model. Due to common hardware limitations, it is very difficult to get a practical speedup from unstructured compression.

In this paper, we design a novel structured sparse convolution (SSC) filter for convolutional layers,  requiring significantly fewer parameters compared to standard convolution. The proposed filter leverages the inherent spatial properties in the images. The commonly used deep convolutional architectures, when coupled with SSC, outperform other state-of-the-art methods that do pruning at initialization. Unlike typical pruning approaches, the proposed architecture is sparse by design and does not require multiple stages of pruning. The sparsity of the architecture is dataset agnostic and leads to better transfer ability of the model when compared to existing state-of-the-art methods that do pruning at initialization. We also show that the proposed filter has implicit orthogonality that ensures minimum filter redundancy at each layer. Additionally, we show that the proposed filter can be viewed as a generalization of existing efficient convolutional filters used in group-wise convolution (GWC)~\cite{xie2017aggregated}, point-wise convolution (PWC)~\cite{Szegedy2015deeper}, and depth-wise convolution (DWC)~\cite{Vanhoucke2014talk}. Extensive experiments and ablation studies on standard benchmarks depict the efficacy of the proposed filter. Moreover, we further compress existing efficient models such as MobileNetv2~\cite{sandler2018mobilenetv2} and ShuffleNetv2~\cite{ma2018shufflenet} while achieving performance comparable to the original models. 
\section{Methods}
\label{sec:proposed_approach}
\begin{figure}[!t]
	\centering
	\includegraphics[scale=1.2]{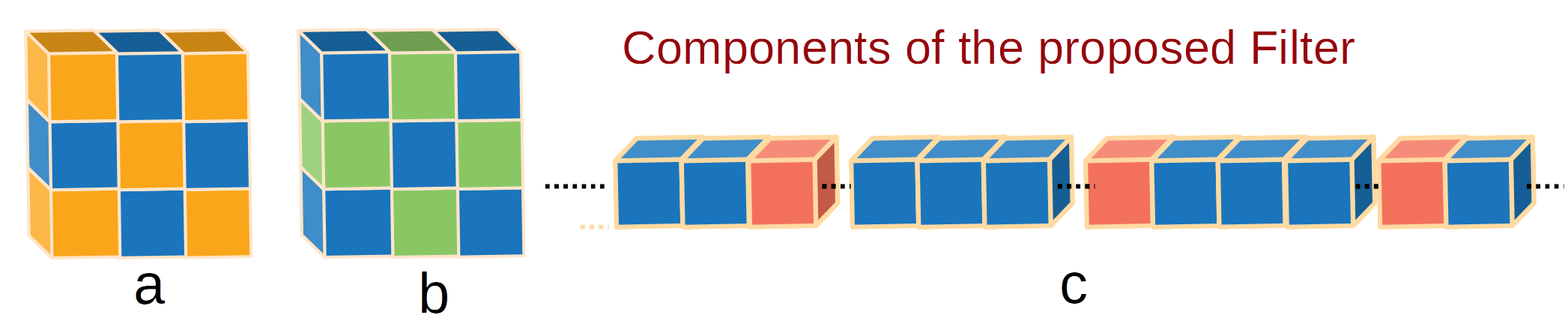}
	\caption{The three basic components that are used in the proposed SSC filter. Blue blocks indicates a zero-weight location. The red, orange and green blocks show the active weight location in three different type of kernels.}
	\label{fig:basic}
\end{figure}

\noindent We propose Structured Sparse Convolution (SSC) filter, which is composed of layered spatially sparse $K \times K$ and $1\times 1$ kernels. Unlike the typical CNN filters that have a kernel of fixed size, the SSC filter has three type of kernels, as shown in Figure~\ref{fig:basic}. The heterogeneous nature of the kernels are designed to have varying receptive fields that can capture different features in the input. As shown in Section~\ref{sec:orthogonal}, heterogeneity in kernels allows the neural network layer to accumulate information from different spatial locations in the feature map while significantly reducing redundancy in the number of parameters.
Consider layer $l$ of a model with an input ($h_{l-1}$) of size $i_{l-1} \times i_{l-1} \times M$, where $i_{l-1}$ corresponds to the spatial dimension (width and height) and $M$ denotes the number of channels of the input. Assume that layer $l$ has $N$ filters, resulting in an output feature map $h_{l}$ of size $i_l \times i_l \times N$.
We represent the computational and memory cost at the $l^{th}$ layer using the number of floating-point operations ($F_l$) and the number of parameters ($P_l$), respectively. The computational and memory cost associated with a standard convolutional layer with a $K \times K$ kernel is the following:
\begin{align}
\small
F_l &= i_l^2 \times N \times (K^2M) , \label{eq:nf_std}\\
P_l &= N \times (K^2M ) , \label{eq:np_std}
\end{align}
where $(K^2M)$ represents the number of total parameters from all $M$ channel-specific kernels. As is evident from (\ref{eq:nf_std}) and (\ref{eq:np_std}), reducing $(K^2M)$ directly reduces both the number of parameters and the computational cost of the model. This is indeed what our proposed method (SSC) achieves -- we design two types of sparse kernels, which form the basic components of SSC.
\begin{figure}[!t]
	\centering
	\includegraphics[scale=0.4]{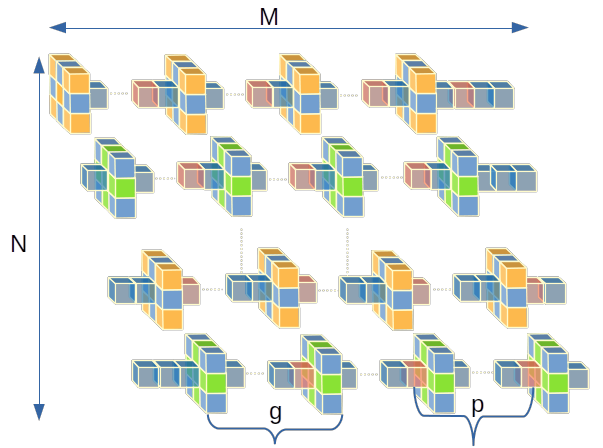}
	\caption{The proposed convolutional layer with $N$ SSC filters. Blue blocks denote the zero-weight locations in $3\times 3$ and $1\times1$ kernels, while other colors show active weights.}
	\label{fig:filter}
\end{figure}
\par
\textbf{Odd/Even $\mathbf{K \times K}$ kernel}: The two types of $K \times K$ kernels differ in terms of the location of the enforced sparsity. Considering $S \in R^{K^2}$ to be the flattened version of the $K \times K$ 2D kernel, we define the odd kernel as:
\begin{equation}
\small
\begin{cases} 
S[i] = 0 & i \in \{2p \enspace | \enspace 0 < 2p < K^2, p\in \mathbb{N} \} \\
S[i] = w_i & i \in \{2p+1 \enspace | \enspace 0 < 2p+1 < K^2, p\in \mathbb{N}\}
\end{cases} ,
\end{equation}
The even kernel is defined in a similar fashion, where the kernel is zero at odd coordinates and non-zero at even coordinates of the filter. Figure~\ref{fig:basic} (a-b) illustrates the odd and even kernel, respectively, when $K=3$. These kernels replace the standard $K \times K$ kernels used in a convolutional layer. 
\par
\textbf{SSC Filter}: A convolutional layer having $N$ SSC filters is shown in Figure~\ref{fig:filter}. An SSC filter is referred to as odd (or even) filter, if it only contains odd (or even) kernels. For each convolutional layer, an equal number of odd and even filters are used. We make the following modifications to a standard convolution filter having $M$, $K\times K$ kernels:

\begin{enumerate}[leftmargin=4mm]
	\setlength{\itemsep}{0.0em}
	\item Among the $M$ different kernels, we replace each kernel at the $k*g$ location with an odd/even kernel, where $g$ is a hyperparameter such that $0 < g<M$ and $k \in \{n \in \sN \enspace | \enspace 0 < k*g \leq M\}$. Note that each filter has only one type (odd/even) of kernel. Each of the $N$ filters have $M/g$ such kernels. The computational cost ($F_{sg}$) and the memory cost ($P_{sg}$) for all the odd/even kernels in a filter:
	\begin{align}
	F_{sg} &= i_l^2 \times N \times \frac{(K^2 - c)M}{g}, \label{eq:33filter} \\
	P_{sg} &= N \times \frac{(K^2-c)M}{g}, \label{eq:33filterpara} \\
	c  &= \Bigg\{\begin{array}{ll}
	\ceil{\frac{K^2}{2}} & \text{Odd kernel}\\
	K^2 - \ceil{\frac{K^2}{2}} & \text{Even kernel}
	\end{array} ,
	\end{align}
	where $c$ represents the number of zeros in the kernel and $\ceil{.}$ denotes the ceiling function.
	\item Out of the remaining $M(1 - 1/g)$ kernel locations in the filter, we place a $1 \times 1$ kernel at a fixed interval of $p$ as shown in Figure~\ref{fig:basic} (c). Each of the $N$ filters have $M(1-1/g)/p$ $1 \times 1$ kernels. The computational and memory cost of these $1 \times 1$ kernels can be defined as:
	\begin{align}
	F_{sp} &= i_l^2 \times N \times \frac{M(1-1/g)}{p} \label{eq:11filter} , \\
	P_{sp} &= N \times \frac{M(1-1/g)}{p} \label{eq:11filterpara} .
	\end{align}
	\item The SSC filter is empty at the remaining $M(1-1/p)(1-1/g)$ locations, causing the filter to ignore the corresponding feature maps (input channels). Note that while a particular filter may not act on certain input features, other SSC filters of the convolutional layer will. This is enforced by the shifting procedure introduced below.
\end{enumerate}

\textbf{Shift operation:} If we naively use SSC filters in a convolutional layer, there will be a loss of information, as all $N$ filters will ignore the same input feature maps. To ensure that each SSC filter attends to a different set of feature maps, we shift the location of all kernels ($K \times K$ and $1\times1$) by $(n\text{ mod }q)$ at initialization\footnote{The shift operation is applied only once before training begins.}, where $n \in \{1,..., N\}$ denotes the index of the filter and $q := max(g,p)$. The shift operation across $N$ filters can be visualized in Figure~\ref{fig:filter}. We can divide the $N$ filters into sets of disjoint filters such that all the filters in a particular set attend to distinct input feature maps. Formally, let the collection of sets be defined as:
\begin{align}
\mathcal{Q} := \{ (0, q), [q, 2q), \dots, [N - (N\text{ mod }q), N)\} , \label{eq:Q-set}
\end{align}
where $[a,a+q)$ denotes the set of filters $a$ through $a+q-1$. Then $\forall f, f' \in [a,a+q)$, $f$ and $f'$ tend to disjoint input feature maps if $f \neq f'$. Moreover, $f$ and $f'$ are ``near-orthogonal'' ($f^Tf' \approx 0$), since they attend to non-overlapping regions of the input feature maps. As discussed in Section~\ref{sec:orthogonal}, the orthogonal property of a layer is of independent interest and allows the network to learn uncorrelated filters. Note that the design of the SSC filter induces structural sparsity as the sparse region is predetermined and fixed, which is in contrast to the unstructured pruning method \cite{frankle2018lottery, lee2018snip, wang2020picking}.

We can quantify the total reduction in the number of floating-point operations ($R_F$) and the number of parameters ($R_p$) with respect to the standard convolutional layer:
\begin{align}
R_F =& \left(1-\frac{F_{sg}+F_{sp}}{F_l}\right)\times100\% \\
=& \left(1-\frac{(1-c/K^2)}{g}-\frac{(1-1/g)}{K^2p}\right)\times100\% ,
\label{eq:flopreduction}\\
R_p =& \left(1-\frac{P_{sg}+P_{sp}}{P_l}\right)\times100\%\\
= & \left(1-\frac{(1-c/K^2)}{g}-\frac{(1-1/g)}{K^2p}\right)\times100\% ,
\label{eq:parareduction}
\end{align}
where $0<g$ and $p<M$. The hyperparameters $p$ and $g$ are set to achieve the desired sparsity in the architectures; we use $R_p$ as guiding principle behind choosing $p$ and $g$. One can also achieve a desired reduction in floating-point operations ($R_F$) to determine the corresponding hyperparameters. However, in our experiments we consider sparsity constraints only.

\subsection{Implicit Orthogonality}
\noindent
\label{sec:orthogonal}
Recent work~\cite{shang2016understanding, xie2017all} shows that deep convolutional networks learn correlated filters in overparameterized regimes. This implies filter redundancy and correlated feature maps when working with deep architectures. The issue of correlation across multiple filters in the convolutional layer has been addressed by incorporating an explicit orthogonality constraint to the filter of each layer~\cite{bansal2018can,wang2020orthogonal}. Consider a 2D matrix $W \in \mathbb{R}^{J \times N}$ containing all the filters: $W=[f_1,f_2,\dots, f_N]$, where $f_n\in \mathbb{R}^{J}$ is the vector containing all the parameters in the $n^{th}$ filter, and $J = K^2M$ for a standard convolutional layer. The soft-orthogonality (SO) constraint on a layer $l$ with the corresponding 2D matrix $W_l$ is defined as: 
\begin{equation}
L_{SO}=\lambda||W_l^TW_l-I||_F^2 , \label{eq:softorth}
\end{equation}
\noindent where $I\in \mathbb{R}^{N\times N}$ is the identity matrix, $\lambda$ controls the degree of orthogonality and $||.||_F^2$ is the Frobenius norm. However, note that the columns of $W$ can only be mutually orthogonal if $W_l$ is an undercomplete or a square matrix ($J \geq N$), which may not be the case in practice. For overcomplete settings ($J<N$), $W_l^TW_l$ can be far from the identity, since the rank of $W_l$ will be upper-bounded by $J$. This makes the optimization in (\ref{eq:softorth}) a biased objective. To alleviate this issue, double soft orthogonality (DSO) regularization has been adopted~\cite{bansal2018can}, which covers both overcomplete and undercomplete cases:
\begin{equation}
L_{DSO}=\lambda\left(||W_l^TW_l-I||_F^2+||W_lW_l^t-I||_F^2\right) .
\label{eq:dualsoftorth}
\end{equation}
Both the constraints $L_{SO}$ and $L_{DSO}$ are commonly used regularization techniques to encourage filter diversity. While the above regularization methods provide a reasonable result by decreasing the correlation across filters, they have several limitations: ($1$) The objectives in (\ref{eq:softorth}-\ref{eq:dualsoftorth}) are computationally expensive and have to be computed over all the layers of the convolutional network. ($2$) The above objectives do not enforce uncorrelated filters but only encourage filter diversity, making such regularization dependent on the dataset complexity. For example, a dataset with fewer training examples may lead to a large number of redundant filters compared to a complex dataset having many training examples.

In contrast, the SSC filters induce group-wise ``near-orthogonality''. In particular, for each set $[a, a+q)$ as defined in (\ref{eq:Q-set}), the $q$ filters are implicitly pairwise orthogonal as they operate at non-overlapping regions of the input feature maps. In particular, the odd/even kernels combined with the shift operation lead to implicit orthogonality. This can be easily visualized from Figure~\ref{fig:filter}. The only source of potential redundancy in SSC filters could arise from the $1\times1$ kernels, whose receptive field may overlap with odd/even kernels. However, as we show in Section~\ref{sec:parameter_redundancy}, the average pairwise correlation is significantly lower compared to the correlation in a standard convolutional layer. Moreover, a key advantage of using SSC filters is that we do not require expensive explicit layer-wise regularization objectives (\ref{eq:softorth}) and (\ref{eq:dualsoftorth}). The reduction in the correlation among filters is a byproduct of the sparsity enforced in the SSC filters, which is complemented by low computational and memory costs as shown in (\ref{eq:flopreduction}) and (\ref{eq:parareduction}). We provide a comparison of the correlation obtained from the proposed filter with the correlation obtained in a standard model with SO and DSO constraints in the experiments.
\subsection{Connection to other efficient filters}
\noindent
\label{sec:connections}
The existing efficient filters used for groupwise convolution (GWC), depthwise convolution (DWC) and pointwise convolution (PWC) can be seen as special cases of the proposed SSC filters. The GWC, DWC and PWC have proven to be essential components for the state-of-the-art architectures designed for low-end computational devices, such as mobile devices~\cite{sandler2018mobilenetv2, ma2018shufflenet}. 

For convenience, we denote the hyperparameter setting $g=0$ to the case where no odd/even kernel is included in the convolutional layer, $i.e.$, there is no sparsity within the kernel. Similarly, $p=0$ denotes the setting where no $1\times1$ kernel is included. If we set $p=0$, the SSC filters can be used for GWC, where each filter will be of size $K \times K \times M/g$ and the kernels operate on a group of feature maps that are separated by an interval of $g$. The DWC operation can be achieved by using a SSC filter with $p=0$ and $g=M$, where each filter will act on a single input feature map. The PWC operation can be achieved with the SSC filter by setting $g=0$ and $p=1$. While the SSC operation provides a generic framework that can be reduced to GWC, DWC and PWC, the SSC filters are inherently sparse. One can get the standard (non-sparse) version of these efficient filters by using the standard $K \times K$ kernels instead of odd/even kernels defined in Section~\ref{sec:proposed_approach}. 

Current state-of-the-art efficient architectures MobileNetv2~\cite{sandler2018mobilenetv2} and ShuffleNetv2~\cite{ma2018shufflenet} reduce the number of parameters by using a two-layer sequential step: DWC followed by a PWC operation. While these operations are less computationally expensive compared to the standard convolutional layer, the latency of the model increases due to the two-layer sequential step. In contrast, the proposed SSC filter can achieve the two operations (GWC/DWC followed by PWC) using a single layer without increasing the number of parameters. We can perform a composition of GWC and PWC by using $p=1,g > 0$. Similarly, a composition of DWC and PWC can be done using $p=1$ and $g=M$. In  Table \ref{table:efficient_arch} we show the results of above composition. 


\section{Related Work}
\noindent The area of model compression has seen immense progress in recent years. Below, we highlight some recently proposed techniques for learning sparse neural networks. We divide the model compression and pruning techniques into three categories based on the computational cost involved.

\textbf{Pruning After Training} (PAT)~\cite{srinivas2015data,han2015deep,han2015learning,frankle2018lottery,frankle2019lottery,wen2016learning, yoon2017combined,singh2019play,singh2020leveraging} is the most widely used pruning method. \cite{han2015learning} proposed pruning the weight parameters using iterative thresholding. Knowledge distillation based methods~\cite{sanh2019distilbert,jiao2019tinybert,chen2020distilling} attempt to train a compressed student model that mimics the behaviour of the full-sized original model. \cite{jaderberg2014} and \cite{denton2014exploiting} proposed low-rank approximation of the weight tensors to reduce the memory and time complexity at training and testing time. These approaches tend to accumulate errors in the prediction when multiple layers are compressed.
PAT-based pruning methods are typically costly and time-consuming, since they require pretraining of the original overparameterized model.

\textbf{Dynamic Pruning} (DP)~\cite{lin2020dynamic,wang2020dynamic,mostafa2019parameter,mocanu2018scalable} involves pruning and training the model simultaneously; as the training progresses the model size decreases. Soft Filter Pruning (SFP)~\cite{he2018soft} prunes the filter after each epoch, but updates the pruned filters when training the model. Deep Rewiring (DeepR)~\cite{bellec2018deep} and Dynamic Sparse~\cite{mostafa2019parameter} prune and re-grow the architecture periodically, but are more computationally expensive when pruning large networks. Sparse momentum (SM)~\cite{dettmers2019sparse} also follows the prune and re-grow approach, however, it uses smoothed gradients to accelerate the training. These approaches can be trained more efficiently compared to PAT-based methods, since a model trained with DP shrinks with training.

\textbf{Pruning Before Training (PBT)} is the most challenging and yet the most practically useful setting among the three categories. Despite its importance, there have been only a few attempts that explore this setting. These approaches find a small sub-network before the training begins and thus require less computational resources and training time. While there have been some attempts to prune the deep neural network before the training~\cite{lee2018snip, Lee2020A, wang2020picking,wimmer2022interspace,singh2019hetconv}, they still require multiple forward/backward passes through the full model to detect unimportant connections in the network. Single-Shot Network Pruning (SNIP)~\cite{lee2018snip} tries to identify a sparse sub-network by solving a minimization problem that preserves the loss of the pruned network at initialization. \cite{Lee2020A} studied the pruning problem from a signal propagation perspective and proposed using an orthogonal initialization to ensure faithful signal propagation. Recently, \cite{wang2020picking} proposed Gradient Signal Preservation (GraSP) to prune the network at initialization by preserving gradient flow. Interspace Pruning (IP)~\cite{wimmer2022interspace} was recently proposed to overcome the bias introduced in PBT methods, which as a result improves the generalization of existing unstructured pruning methods (including SNIP and GraSP). Although these methods are developed to find a sub-network at initialization, they still require optimizing the original overparameterized model using the training dataset, which can be expensive for low-end devices. Moreover, the sub-networks found are specific to a particular dataset, hindering knowledge transfer across multiple tasks. Synaptic Flow (SynFlow)~\cite{tanaka2020pruning} prunes weights using a information throughput criterion to find a sparse network. Our proposed method also falls into the PBT category, where the sparse network is identified at the initialization. In contrast to previous methods, the proposed method is sparse by design and does not require solving an optimization problem to find a task-specific sub-network. 

\section{Experiments}
\begin{table*}[!th]
\centering
\caption{\small Test accuracy of pruned VGG19 and ResNet32 on CIFAR-10 and CIFAR-100 datasets. The bold number is the higher one between the accuracy of GraSP, SNIP and SSC.}
\vspace{0.2em}
\resizebox{0.85\textwidth}{!}{%
\begin{tabular}{l||ccc|ccc}
\toprule
\textbf{Dataset}     & \multicolumn{3}{c|}{CIFAR-10} & \multicolumn{3}{c}{CIFAR-100}  \\ 
\toprule
Pruning ratio     & 90\%      & 95\%     & 98\%      & 90\%      & 95\%     & 98\%       \\ 
\midrule
\textbf{VGG19}~(Unpruned)
&  94.23 & - & - & 74.16 & - & - \\
OBD~\cite{lecun1990optimal}& 93.74 & 93.58 & 93.49 & 73.83 & 71.98 & 67.79 \\
MLPrune~\cite{zeng2019mlprune}& 93.83 & 93.69  &  93.49& 73.79  & 73.07  &  71.69\\
LT~(original initialization)& 93.51 & 92.92 & 92.34 & 72.78 & 71.44 & 68.95 \\
LT~(reset to epoch 5)& 93.82 & 93.61  & 93.09 & 74.06 & 72.87  & 70.55 \\
\arrayrulecolor{gray}\cmidrule(lr){1-7}\arrayrulecolor{black}
DSR~\cite{mostafa2019parameter}& 93.75 & 93.86  & 93.13 & 72.31 & 71.98  & 70.70 \\
SET~\cite{mocanu2018scalable}& 92.46 & 91.73  & 89.18 & 72.36 & 69.81  & 65.94 \\
Deep-R~\cite{bellec2018deep}& 90.81 & 89.59  & 86.77 & 66.83 & 63.46  & 59.58 \\
\arrayrulecolor{gray}\cmidrule(lr){1-7}\arrayrulecolor{black}
SNIP~\cite{lee2018snip} &{93.63$\pm$0.06} & {93.43$\pm$0.20} & 92.05$\pm$0.28&{72.84$\pm$0.22}	& {71.83$\pm$0.23} & 58.46$\pm$1.10\\
GraSP~\cite{wang2020picking} &93.30$\pm$0.14& 93.04$\pm$0.18 & {92.19$\pm$0.12}&71.95$\pm$0.18 &71.23$\pm$0.12 &{68.90$\pm$0.47}\\
SynFlow~\cite{tanaka2020pruning} &92.99$\pm$0.18& 92.23$\pm$0.15 & {91.01$\pm$0.17}&--.-- &--.-- &--.--\\
IP-SynFlow~\cite{wimmer2022interspace} &93.17$\pm$0.20& 92.46$\pm$0.14 & {92.11$\pm$0.24}&--.-- &--.-- &--.--\\
\textbf{SSC (Ours)} & \textbf{93.68$\pm$0.11}& \textbf{93.45$\pm$0.17} & {91.16$\pm$0.20}&71.65$\pm$0.18 &70.71$\pm$0.21 &\textbf{68.95$\pm$0.44}
\\\bottomrule\bottomrule
\end{tabular}
}\\
\resizebox{0.85\textwidth}{!}{%
\begin{tabular}{l||ccc| ccc}
\textbf{ResNet32}~(Unpruned)
& 94.80 & - & - & 74.64 & - & - \\
OBD~\cite{lecun1990optimal}& 94.17 & 93.29 & 90.31 & 71.96 & 68.73 & 60.65 \\
MLPrune~\cite{zeng2019mlprune}& 94.21  & 93.02  & 89.65& 72.34  & 67.58  &  59.02 \\
LT~(original initialization)& 92.31 & 91.06  & 88.78 & 68.99 & 65.02  & 57.37 \\
LT~(reset to epoch 5)& 93.97 & 92.46  &  89.18& 71.43 & 67.28  &  58.95\\
\arrayrulecolor{gray}\cmidrule(lr){1-7}\arrayrulecolor{black}
DSR~\cite{mostafa2019parameter}& 92.97 & 91.61  & 88.46 & 69.63 & 68.20  & 61.24 \\
SET~\cite{mocanu2018scalable}& 92.30 & 90.76  & 88.29 & 69.66 & 67.41  & 62.25 \\
Deep-R~\cite{bellec2018deep}& 91.62 & 89.84  & 86.45 & 66.78 & 63.90  & 58.47 \\
\arrayrulecolor{gray}\cmidrule(lr){1-7}\arrayrulecolor{black}
SNIP~\cite{lee2018snip} & {92.59$\pm$0.10} & 91.01$\pm$0.21 &87.51$\pm$0.31& 68.89$\pm$0.45 & 65.22$\pm$0.69 &54.81$\pm$1.43\\
GraSP~\cite{wang2020picking}&92.38$\pm$0.21 &{91.39$\pm$0.25} &{88.81$\pm$0.14}&{69.24$\pm$0.24} &{66.50$\pm$0.11} &{58.43$\pm$0.43}\\ 
\textbf{SSC (Ours)}
&\textbf{93.29$\pm$0.20} &\textbf{91.66$\pm$0.13} &88.39$\pm$0.24
&\textbf{71.13$\pm$0.12} &\textbf{67.31$\pm$0.15} &\textbf{63.01$\pm$0.31}
\\ 
\bottomrule
\end{tabular}
}
\label{table:classification_cifar}
\end{table*}
\label{sec:experiments}
\noindent We demonstrate the efficacy and efficiency of our proposed filter with an extensive set of experiments. First, we evaluate the performance of commonly used deep convolutional networks (ResNet-32/50 and VGG-19~\cite{He2016, simonyan2014very}) with SSC filters on four classification benchmarks: CIFAR-10/100~\cite{krizhevsky2009learning}, Tiny-ImageNet~\cite{wu2017tiny} and ImageNet~\cite{deng2009imagenet}. We show that the proposed SSC filters achieve state-of-the-art accuracy in most settings. We also apply SSC filters to existing state-of-the-art ``efficient" architectures: MobileNetV2~\cite{sandler2018mobilenetv2} and ShuffleNetV2~\cite{ma2018shufflenet}. In Section~\ref{sec:compress_efficient}, we show that these architectures can be further compressed by 47-48\% with SSC, while achieving high accuracy on the CIFAR-10 benchmark.
\par
Next, we conduct experiments to analyze the properties of SSC filters. Specifically, we find that SSC leads to significantly lower layer-wise filter correlation, implying fewer redundant filters compared to alternative methods that use explicit regularizers. We also test the robustness of SSC to overfitting when trained with limited training data. Finally, we evaluate the ability of SSC filters in transfer learning.

\begin{table*}[!ht]
\centering
\caption{Test accuracy of pruned VGG19 and ResNet32 on Tiny-ImageNet dataset. The bold number is the higher one between the accuracy of GraSP and that of SNIP.}
\vspace{0.2em}
\resizebox{0.85\textwidth}{!}{%
\begin{tabular}{l|ccc| ccc}
\toprule
\textbf{Network}     & \multicolumn{3}{c}{VGG19}   &     \multicolumn{3}{c}{ResNet32}\\ 
\midrule
Pruning ratio    & 90\%      & 95\%     & 98\% & 85\% & 90\%   & 95\%   \\ 
\midrule
\textbf{VGG19/ResNet32}~(Unpruned)& 61.38 & - & - &  62.89 & - & - \\
OBD~\cite{lecun1990optimal}& 61.21 & 60.49 &  54.98& 58.55 & 56.80 & 51.00  \\
MLPrune~\cite{zeng2019mlprune} & 60.23 & 59.23 & 55.55& 58.86 & 57.62 & 51.70  \\
LT~(original initialization)& 60.32 & 59.48  & 55.12   & 56.52 & 54.27  & 49.47\\
LT~(reset to epoch 5)& 61.19 & 60.57   &  56.18& 60.31 & 57.77  &  51.21\\
\arrayrulecolor{gray}\cmidrule(lr){1-7}\arrayrulecolor{black}
DSR~\cite{mostafa2019parameter}& 62.43 & 59.81  & 58.36 & 57.08 & 57.19  & 56.08 \\
SET~\cite{mocanu2018scalable}& 62.49 & 59.42 & 56.22 & 57.02 & 56.92 & 56.18 \\
Deep-R~\cite{bellec2018deep}
& 55.64 &  52.93 & 49.32& 53.29 & 52.62  & 52.00 \\
\arrayrulecolor{gray}\cmidrule(lr){1-7}\arrayrulecolor{black}
SNIP~\cite{lee2018snip} & {61.02$\pm$0.41} & 59.27$\pm$0.39 & 48.95$\pm$1.73& 56.33$\pm$0.24 & 55.43$\pm$0.14 & 49.57$\pm$0.44\\
GraSP~\cite{wang2020picking}
& 60.76$\pm$0.23 & {59.50$\pm$0.33} & {57.28$\pm$0.34}& {57.25$\pm$0.11} & {55.53$\pm$0.11} & {51.34$\pm$0.29} \\
\textbf{SSC (Ours)}
& \textbf{61.53$\pm$0.25} & \textbf{59.72$\pm$0.23} & {56.81$\pm$0.39}
& \textbf{57.92$\pm$0.16} & \textbf{57.36$\pm$0.19} & \textbf{53.70$\pm$0.21} \\
\bottomrule
\end{tabular}
}
\label{table:classification_tiny_imagenet}
\end{table*}
For all the experiments, we follow the training setup (optimizer, learning rate, training epochs) used in \cite{wang2020picking}. We include all hyperparameters in the supplementary material. In addition to the aforementioned experiments, we carry out extensive ablations on choosing the hyperparameters $g$ and $p$. In general, the hyperparameters $p$ and $g$ are set to achieve the desired sparsity in the architectures; we use $R_p$ (as defined in \ref{eq:parareduction}) as our guiding principle behind choosing $p$ and $g$. For brevity, we defer the ablation studies and further discussion to the supplementary.\\

\noindent \textbf{Baselines:} We include a number of baselines to compare the performance of the proposed approach. The baselines include methods that prune the architecture after complete training, such as OBD~\cite{hassibi1993optimal}, MLPrune~\cite{zeng2019mlprune} and LT~\cite{frankle2018lottery}. We also consider methods that do dynamic pruning: DSR~\cite{mostafa2019parameter}, SET~\cite{mocanu2018scalable} and Deep-R~\cite{bellec2018deep}. From the class of methods that do pruning before training (PBT), we consider SNIP~\cite{lee2018snip}, GraSP~\cite{wang2020picking}, and SynFlow~\cite{tanaka2020pruning}. All the above baselines, except PBT methods, have an added advantage of training a large overparameterized network compared to our proposed method that trains on a highly-sparse architecture from initialization. While we report performance of all the above baselines, a fair comparison of our method can only be done with SynFlow, SNIP and GraSP. We also consider the improvements introduced with IP~\cite{wimmer2022interspace} for PBT methods.

\subsection{Performance on Classification benchmarks}
\noindent We evaluate the proposed method on the CIFAR10 and CIFAR100 classification benchmark by training commonly used architectures (VGG-19 and ResNet-32) with the standard convolution filters replaced by the proposed SSC filters. We report the test accuracy using VGG-19 and ResNet-32 under three pruning ratios, namely 90\%, 95\% and 98\% in Table~\ref{table:classification_cifar}. We observed that SSC performs better than SNIP and GraSP in 8 of the 12 different settings considered, especially when we use ResNet32 (outperforming in 5 of 6 settings). Note that even in the settings where SSC is inferior to SNIP or GraSP, the performance is highly competitive to the best method. Moreover, in the extreme setting of 98\% sparsity on the CIFAR-100 dataset, SSC outperforms the next best method using ResNet32 with a 7.8\% relative improvement.

Tiny-ImageNet is a medium-scale dataset containing images of 200 classes from the ImageNet. Again, we chose VGG-19 and ResNet-32 as the base architectures with varying pruning ratio. The results are reported in the Table~\ref{table:classification_tiny_imagenet}. Once again, we observe that our approach outperforms the baselines methods SNIP and GraSP in 5 of the 6 settings. In the extreme setting of 95\% sparsity in ResNet-32, we observe that the proposed method shows a relative improvement of 4.6\% (absolute improvement of 2.36\%) over GraSP.

We also conduct a large scale experiment on the ImageNet dataset using the ResNet-50 architecture. Results are shown in Table~\ref{table:ResNet-50_imagenet_classification}. The model is trained under two pruning ratios: 60\% and 80\%. We report the top-1 and top-5 accuracy using SNIP, GraSP and our proposed method. Despite the added advantage in SNIP and GraSP that use memory intensive ``foresight pruning'' on the large ResNet-50 architecture, our proposed SSC performs comparably. For the 60\% pruning ratio, the top-1 accuracy of SSC only lags behind by 0.26\% and the top-5 by 0.14\% in absolute difference. This makes SSC based deep convolutional models appealing for devices that cannot train large models as SSC does not require iterative pruning on a large model. We also conducted the ImageNet experiment over the ResNet18 architecture and the results are shown in the Table~\ref{table:ResNet-18_imagenet_classification}. In spite of the additional overhead used in baseline methods, we find SSC to be better or comparable to the baselines. 
\begin{table}
\centering
\caption{The test accuracy using ResNet-50 architecture on the
ImageNet benchmark with pruning ratios 60\% and 80\%. The baselines SNIP and GraSP have additional overhead compared to SSC. Nevertheless, SSC performs comparably/better than SNIP and GraSP.}
\vspace{0.5em}
\label{table:ResNet-50_imagenet_classification}
\resizebox{0.42\textwidth}{!}{%
\begin{tabular}{lccc c}
\toprule
\textbf{Pruning ratios} & \multicolumn{2}{c}{60\%}   & \multicolumn{2}{c}{80\%} \\
\midrule
Accuracy & top-1 & top-5 & top-1 & top5 \\
\arrayrulecolor{gray}\cmidrule(lr){1-5}\arrayrulecolor{black}
\textbf{ResNet-50}~(Unpruned) & 75.70 & 92.81 & - & - \\ 
SNIP~\cite{lee2018snip} & 73.95  & {91.97}  & 69.67  & 89.24 \\ 
GraSP~\cite{wang2020picking} & {74.02} & 91.86  & {72.06} & {90.82} \\ 
\textbf{SSC (Ours)} & 73.76 & 91.83  & 71.27 & {90.32} \\ 
\bottomrule
\end{tabular}
}
\end{table}

\begin{table}
\centering
\caption{The test accuracy using ResNet-18 architecture on the ImageNet benchmark with pruning ratios 60\% and 75\%. All the baselines have additional overhead compared to SSC.}
\vspace{0.5em}
\label{table:ResNet-18_imagenet_classification}
\resizebox{0.44\textwidth}{!}{%
\begin{tabular}{lcccc} 
\toprule
\textbf{Pruning ratios} & \multicolumn{2}{c}{60\%}   & \multicolumn{2}{c}{75\%} \\
\midrule
Accuracy & top-1 & top-5 & top-1 & top5 
\\
\arrayrulecolor{gray}\cmidrule(lr){1-5}\arrayrulecolor{black}
\textbf{ResNet-18}~(Unpruned) & 69.77 & 89.07 & - & - \\
SNIP~\cite{lee2018snip} & 66.56  & 87.32  & 64.33  & 85.74 \\
GraSP~\cite{wang2020picking} & 65.87 & 86.71  & 63.57 & 85.34 \\
SynFlow~\cite{tanaka2020pruning} & 66.93 & 87.56  & 63.85 & 85.61 \\
IP-SNIP~\cite{wimmer2022interspace} & 67.29  & 87.65  & 64.93  & 86.18 \\
IP-GraSP~\cite{wimmer2022interspace} & 66.92 & 87.48  & 64.94 & 86.20 \\
IP-SynFlow~\cite{wimmer2022interspace} & 67.31 & 87.85  & 64.63 & 86.11 \\
\textbf{SSC (Ours)} & 67.55 & 87.89  & 64.86 & 85.83 \\
\bottomrule
\end{tabular}
}
\end{table}

\subsection{Compression of the efficient Architecture}
\begin{table}[!t]
\centering
\caption{Compression of MobileNet and ShuffleNet architectures on the CIFAR-10 benchmark. Note that baselines require pruning after training, which is more computationally expensive than SSC.}
\label{table:efficient_arch}
\resizebox{0.44\textwidth}{!}{%
\begin{tabular}{clll} 
\toprule
  & Method   & Param($\%$) & Acc ($\%$)\\
\midrule
\multirow{5}{*}{MobileNetV2} &
Unpruned & 100.0 & 94.3 \\ 
& $L_1$P~\cite{li2016pruning}  & 51.3 & 84.4 \\
& Slimming~\cite{liu2017learning} & 43.8 & 91.5 \\
& AutoSlim~\cite{yu2019autoslim} & 46.0 & 91.9\\
\arrayrulecolor{gray}\cmidrule(lr){2-4}\arrayrulecolor{black}
& \textbf{SSC (Ours)} & \textbf{52.4} & \textbf{92.9}\\
\midrule
\multirow{3}{*}{ShuffleNetV2} & Unpruned & 100.0 & 91.0 \\
& Slimming~\cite{liu2017learning} & 55.8 & 89.5\\
\arrayrulecolor{gray}\cmidrule(lr){2-4}\arrayrulecolor{black}
& \textbf{SSC (Ours)} & 52.2 & \textbf{89.7}\\
\bottomrule
\end{tabular}
}
\vspace{-1em}
\end{table}
\label{sec:compress_efficient}
\noindent We apply the proposed SSC filter to the MobileNetV2~\cite{sandler2018mobilenetv2} and ShuffleNetV2~\cite{ma2018shufflenet} models, which are the state-of-the-art architectures for low-end devices. As discussed in Section~\ref{sec:connections}, we replace the DWC and PWC operations with the proposed SSC operation, allowing us to further compress the efficient architectures~\cite{sandler2018mobilenetv2, ma2018shufflenet}. We set the hyperparameters as follows: $(1)$ For all PWC filters, we fix $p=2$ and $g=0$. This reduces the total number of PWC parameters in the standard architecture by 50\%. $(2)$ For the DWC filters, we set $p=0$ and $g=M$. We do not use the odd/even kernel in the SSC filter as there is only a single kernel in DWC filters. We report the results in Table~\ref{table:efficient_arch}. We consider $L_1P$~\cite{li2016pruning}, Slimming~\cite{liu2017learning} and AutoSlim~\cite{yu2019autoslim} as the baseline models. Unlike the proposed method, all three baselines require a pretrained model to prune the model. Despite this advantage over SSC, the SSC filter shows significant improvement over the baselines for the MobileNetV2 and ShuffleNetV2 architectures. 

\begin{figure*}[!th]
	\centering
    \includegraphics[width=\textwidth]{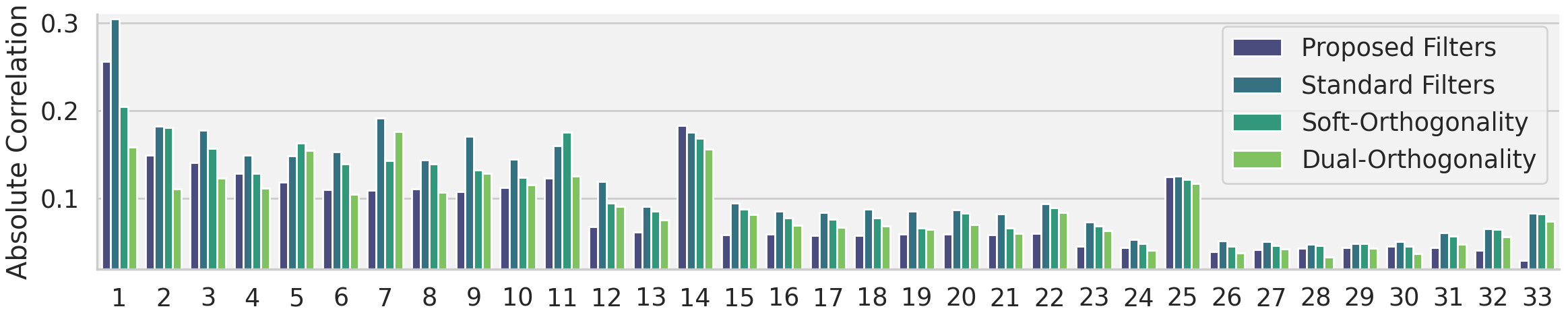}
	\caption{The filter correlation (Y axis) in ResNet-32 (X axis: Layer) for the proposed SSC (10\% parameters) and baselines.}
	\label{fig:corr}
\end{figure*}
\subsection{Parameter Redundancy}
\label{sec:parameter_redundancy}
\noindent In Section~\ref{sec:orthogonal}, we discussed the implicit orthogonal nature of the SSC filters. We compare the average absolute correlation at each layer of the ResNet-32 architecture trained with and without using the SSC filters. The pairwise correlation ($\rho$) between the $i^{th}$ and $j^{th}$ filter can be calculated as: $\rho(f_i,f_j)=\mathbb{E}[(f_i-\mu_i)(f_j-\mu_j)]/(\sigma_i\sigma_j)^{1/2}$, where $f_i \in \mathbb{R}^{K^2M}$ represents the 1D vector of all the parameters in the $i^{th}$ filter and $\mu_i$ is the mean of all the parameters in $f_i$ and $\sigma_i=\sum_{i=1}^{K^2M}(f_i-\mu_i)$. We define the average absolute correlation at layer ${\ell}$: $C^{\ell}=\frac{1}{N^2}\sum_{i=1}^N\sum_{j=1}^N |\rho(f^{\ell}_i,f^{\ell}_j)|$. In Figure~\ref{fig:corr}, we report the correlation measure $C^{\ell}$ at each layer of the ResNet-32 architecture trained on the CIFAR-100 benchmark. We also compare the correlation measures obtained when a standard ResNet-32 model is trained with explicit orthogonal constraints described in (\ref{eq:softorth}) and (\ref{eq:dualsoftorth}).

As seen in Figure~\ref{fig:corr}, the correlation measure of the standard convolutional filters is higher compared to other methods, implying filter redundancy in the standard ResNet-32 architecture. We observe that for 31 layers out of the total 33, the SSC filters have lower correlation measure compared to standard filters. In comparison to methods with explicit regularization constraints, the SSC filters have lower correlation measure for 21 layers. This shows that SSC-based ResNet architecture can utilize parameters more effectively by learning diverse filters compared to the baselines. Moreover, the SSC filters circumvent the under/over-complete issues (Section~\ref{sec:orthogonal}), while avoiding expensive computations (\ref{eq:softorth}-\ref{eq:dualsoftorth}).     
\vspace{-0.7em}
\subsection{Robustness to overfitting}
\begin{figure}
\centering
\includegraphics[height=3cm,width=3.8cm]{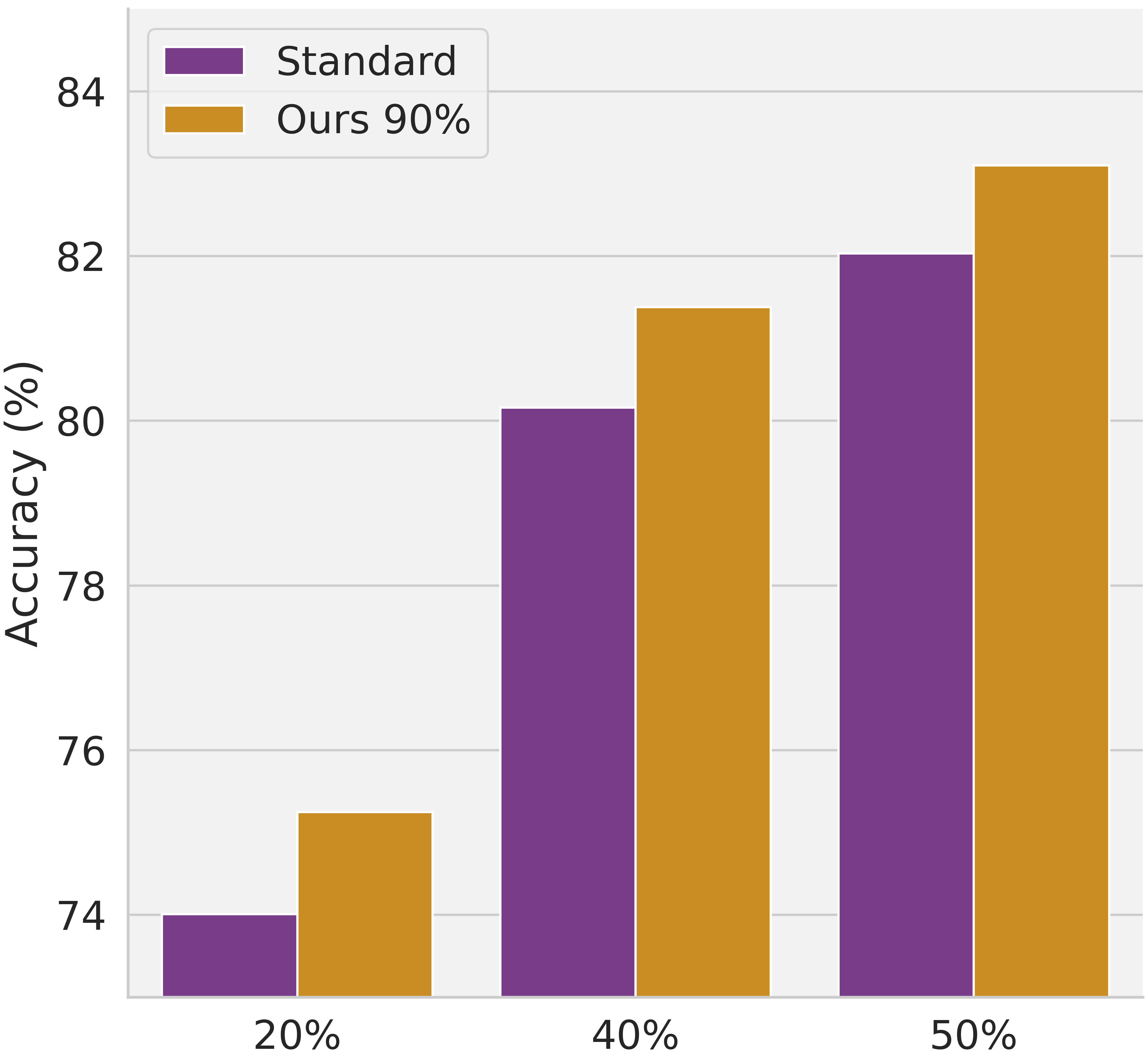}\quad
\includegraphics[height=3cm,width=3.8cm]{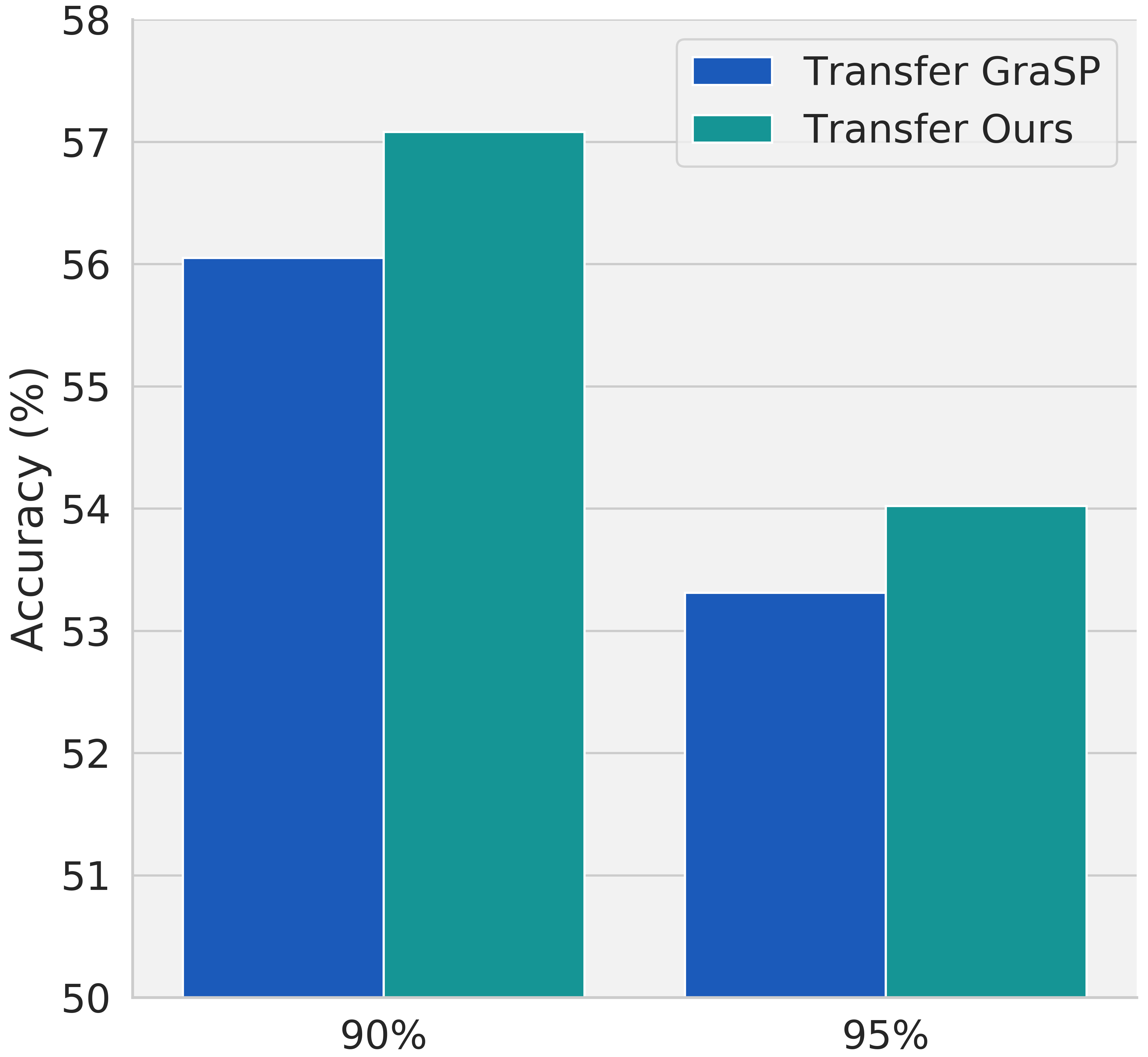}
\vspace{0.5em}
\caption{\textbf{Left:} Accuracy with ResNet-32 on the CIFAR-10 dataset when only 20\%, 40\% and 50\% samples are used. \textbf{Right:} Accuracy of sparse networks when transferred to a different task.}
\label{fig:overfitting}
\end{figure}

\noindent An overparameterized deep learning model is known to overfit in a low-data regime. As shown empirically in Section~\ref{sec:parameter_redundancy}, the proposed SSC filter can learn diverse filters. In this section, we analyze the robustness of the SSC filters when the training data is scarce. To test if the SSC filters are robust to overfitting, we train the ResNet-32 architecture on partial training data. We consider three scenarios that involve learning the model using only 20\%, 40\% and $50\%$ images of the total training set. The performance of the three scenarios is reported in Figure~\ref{fig:overfitting} (Left). The hyperparameters $g$ and $p$ are chosen such that the neural network has only 10\%  of the total parameters in a standard ResNet-32 architecture. As expected, we observe that the model with the SSC filters are more robust to overfitting when compared to the model with the standard convolutional filters. In our experiments, we also observed that the difference in performance becomes more significant when the training data decreases.

\subsection{Transfer Learning}
\label{sec:data_agnostic_model}
\noindent In this section, we study the generalization of the sparse architectures on two different tasks. Recall from Section~\ref{sec:proposed_approach} that the pruning in SSC is governed by choosing the hyperparameters $g$ and $p$; thus, the obtained sparsity in SSC is independent of the training data. Whereas for the baseline methods, the sparse network is found by iterative procedures performed on the training data. We compare the transfer ability of the SSC architecture with the best performing baseline ($i.e.$, GraSP). 
To study the transfer ability across two tasks, we first train GraSP on CIFAR-100 benchmark to prune the original architecture and then transfer (fine-tune) the obtained sparse network on the Tiny-ImageNet dataset. Similarly, for the proposed SSC-based architecture, we first find the sparse network by tuning the hyperparameters $g$ and $p$ on the CIFAR-100 benchmark, and then use the obtained sparse network for training on the Tiny-ImageNet dataset. To keep the complexity of the Tiny-ImageNet task same as CIFAR-100, we only use 100 classes from the Tiny-ImageNet dataset. The performance of the two methods (GraSP and Ours) under pruning ratios of $90\%$ and $95\%$ is shown in Figure~\ref{fig:overfitting} (right). We observe that the SSC-based architectures performs better than GraSP, depicting better transfer ability.

\section{Conclusions}
\noindent We have proposed structured sparse convolutional (SSC) for deep convolutional neural networks. SSC is based on efficient filters composed of novel odd/even kernels and $1 \times 1$ kernels. The proposed kernels leverage the spatial dependencies in the input features to reduce the floating-point operations and the number of parameters in the deep neural network from initialization. Through a series of experiments, we demonstrate the efficacy of the proposed SSC when applied to commonly used deep convolutional networks. A key attribute of SSC filters is that unlike existing approaches, SSC requires no additional pruning during/after training. We also show that the SSC filters generalize other efficient filters (GWC, DWC, and PWC) and demonstrate the applicability of SSC-filters on existing efficient architectures like MobileNet and ShuffleNet.

While this work has demonstrated potential in designing sparse filters without additional steps, our proposed filter has limited practical speedup as existing deep learning libraries do not support efficient operations on structured tensors. However, an efficient CUDA implementation to incorporate such operations for structured sparsity can easily solve this issue. We strongly believe that the proposed method could be highly beneficial to the broader community in training powerful deep systems with only a fraction of computational resources. Regardless, there is still little that we understand about identifying good sparse models that can generalize. We hope that this work will motivate the research community to identify efficient filters that can naturally lead to highly-sparse and effective deep learning models.

{\small
\bibliographystyle{ieee_fullname}
\bibliography{references}
}

\end{document}